\documentclass{article}

\usepackage{PRIMEarxiv}

\usepackage[utf8]{inputenc} 
\usepackage[T1]{fontenc}    
\usepackage{hyperref}       
\usepackage{url}            
\usepackage{booktabs}       
\usepackage{amsfonts}       
\usepackage{nicefrac}       
\usepackage{microtype}      
\usepackage{lipsum}
\usepackage{fancyhdr}       
\usepackage{graphicx}       
\graphicspath{{media/}}     
\usepackage{amsmath}
\usepackage{graphicx}
\graphicspath{{figures/}} 
\usepackage{times,subfigure}
\usepackage{amssymb}
\title{Meta-World Conditional Neural Processes}

\author{
  Suzan Ece Ada, Emre Ugur \\
  Department of Computer Engineering \\
  Bogazici University \\
  Istanbul, Turkey\\
  \texttt{\{ece.ada, emre.ugur\}@boun.edu.tr} \\
}

\begin{document}
\maketitle

\begin{abstract}
We propose Meta-World Conditional Neural Processes (MW-CNP), a conditional world model generator that leverages sample efficiency and scalability of Conditional Neural Processes to enable an agent to sample from its own ``hallucination''. We intend to reduce the agent's interaction with the target environment at test time as much as possible. To reduce the number of samples required at test time, we first obtain a latent representation of the transition dynamics from a single rollout from the test environment with hidden parameters. Then, we obtain rollouts for few-shot learning by interacting with the ``hallucination'' generated by the meta-world model. Using the world model representation from MW-CNP, the meta-RL agent can adapt to an unseen target environment with significantly fewer samples collected from the target environment compared to the baselines. We emphasize that the agent does not have access to the task parameters throughout training and testing, and MW-CNP is trained on offline interaction data logged during meta-training. 
\end{abstract}


\section{Introduction}
Fast adaptation under uncertainty has various applications in the real-world, including autonomous driving, disassembly with manipulation robots, medical diagnosis, and ventilation robots. Because real-world is non-stationary, an agent should be robust to changing environmental dynamics such as failures in the sensors, and actuators \cite{https://doi.org/10.48550/arxiv.2203.01387}. Robustness in these settings requires a diverse set of skills that can be acquired through incremental learning in environments with emergent complexity. In line with this, world model generation has been an integral part of open-ended learning (\cite{DBLP:journals/corr/abs-1901-01753}). Likewise, meta-reinforcement learning \cite{finn2017model} is a promising direction for increasing robustness. However, it is bounded by the number of samples collected from the unknown environment. Increasing sample efficiency is paramount for developing systems that can generate fast responses under uncertainty in the real world. Hence, we propose generating world models from the agent's experience to overcome these problems. 

Research on lifelong learning focuses on learning incrementally and adapting rapidly to unknown tasks from incoming data streams (\cite{PARISI201954}). Challenges associated with lifelong learning include catastrophic forgetting, negative transfer, and resource limitations. Furthermore, data to be processed after deployment and optimal model hyperparameters are not available a priori. 

Meta-reinforcement learning is an enduring area of interest that enables fast adaptation in test tasks. Recent works have integrated meta-learning to offline reinforcement learning (\cite{DBLP:journals/corr/abs-2008-06043}) and lifelong learning (\cite{pmlr-v97-finn19a,DBLP:journals/corr/abs-1812-07671,DBLP:journals/corr/abs-2112-04467}). In open-ended learning, a policy with high representational capacity can be utilized for increasingly challenging environments. In this work, we are interested in few-shot learning in settings where the transition dynamics of the environment change across tasks. More specifically, the agent is trained a the distribution of tasks with varying transition dynamics and expected to adapt to an unseen task with few samples from the target environment. Sampling in the real world is expensive; hence reducing the number of samples used for fast adaptation is an important research direction for sim-to-real RL. 

In this work, we are interested in meta-reinforcement learning settings where the transition function changes across multiple tasks. In particular, each transition function depends on the environment parameters hidden during training and testing. More concretely, imagine a setting where an agent moves to a goal location in different environments parametrized by different force fields. Different force fields push the agent in different directions. Through trial and error, the agent learns to move robustly in environments with different transition dynamics. The agent is required to adapt quickly to a new target environment with hidden environment parameters and reward signals at test time.

We propose Meta-World Conditional Neural Processes(MW-CNP) to generate world models from a few samples collected from the target environment. Our contributions include (1) generating and learning world models with no access to target environment parameters, (2) sample efficient, fast adaptation to the unseen test environment (requires only a single rollout from the unseen target environment at test time compared to 25 rollouts used in No Reward Meta-Learning (NORML) (\cite{yang2019norml}) for finetuning the meta-policy) and (3) utilizing offline datasets of Markov Decision Process(MDP) tuples for training.

\section{Related Work}
\paragraph{Meta Reinforcement Learning}
Meta-learning provides a framework for learning new tasks more efficiently from limited experience by using knowledge from previously learned tasks\cite{Schmidhuber:87long}. Meta-learning framework has demonstrated promising results in regression, classification, and reinforcement learning \cite{finn2017model,yang2019norml}. Our work builds on the meta-reinforcement learning (meta-RL) framework, where tasks share a common structure but differ in transition dynamics. The assumption of a common structure allows few-shot learning in the new task. 

Context-based meta-RL methods assume each task can be represented by a low-dimensional context variable. Prior work on context-based meta-RL learns the task-specific context variable from a set of experiences collected by a control policy. Recurrent models aim to encode the underlying temporal structure of the collected experiences \cite{rl2}. Similarly, in \cite{conf/iclr/MishraR0A18}, memory-augmented models like temporal convolutions and soft attention have been used to learn a contextual representation. In an imitation learning setting, expert demonstrations are encoded into a learned context embedding where policy is conditioned on \cite{conf/nips/DuanASHSSAZ17}. 

In Model-Agnostic Meta-Learning (MAML) \cite{finn2017model}, a gradient-based meta-learning algorithm is used for the aforementioned problems where training and testing tasks are sampled from the distribution. NORML, proposed by \cite{yang2019norml}, is an extension of MAML-RL framework \cite{finn2017model} for settings where the environment dynamics change across tasks instead of the reward function. The goal of NORML is to utilize past experience to quickly adapt to tasks with unknown parameters from a few samples with missing reward signals. While these methods have demonstrated promising results, they fall short in sample efficiency in meta-testing. We instead leverage the generalizable conditional neural processes \cite{cnp} architecture to reduce the number of experience agent needs for adaptation to the target task.  

\paragraph{Partially Observable Markov Decision Process}
In Partially Observable Markov Decision Processes (POMDP) state is augmented by a latent variable $z$ that incorporates unobserved information needed to solve the task. In the Context-based meta-RL setting, solving for the POMDPs become analogous to solving for meta-RL MDPs via task-augmented states \cite{li2020focal}. VARIBAD uses a recurrent neural network (RNN) architecture to encode posterior over ordered transitions. The posterior belief is used to augment the state variable where the policy is conditioned on. Unlike FOCAL \cite{li2020focal} which assumes a deterministic encoder, PEARL \cite{rakelly2019efficient}, ELUE \cite{elue}, and VARIBAD \cite{varibad} uses a stochastic encoder. Latter methods train the inference model by optimizing the evidence lower bound (ELBO)\cite{vae} objective. Although FOCAL successfully clusters transitions in the latent space using an inverse power loss distance metric learning objective, it assumes the tasks are deterministic. Furthermore, FOCAL assumes there exists a one-to-one function from the transitions to the $T \times R$ where $T$ and $R$ denote the task-specific transition function and reward function respectively. 
\paragraph{Conditional Neural Processes}
Conditional Neural Processes \cite{cnp} predict the parameters of a probability distribution conditioned on a permutation invariant prior data and target input. CNPs attempt to represent a family of functions using Bayesian Inference and the high representational capacity of neural networks. Prior approaches have leveraged the robust latent representations obtained by the CNP model in high-dimensional trajectory generation \cite{cnmp}, long-horizon multi-modal trajectory prediction \cite{cnmp,SEKER202222} and sketch generation tasks \cite{GAN-CNMP}.

\section{Preliminaries}
\subsection{Conditional Neural Processes}

The architecture of CNPs consists of parameter sharing encoders and a decoder named the query network. In particular, first a random number of input and true output pairs ${(x_{f^{i}},y^{true}_{f^{i}})}$ are sampled from the function ${f^{i}}\in F$. Each pair is encoded into a latent representation via the encoder networks. Then, average of these representations is obtained for invariance to permutation and the number of inputs. The resulting representation is concatenated with the target input query and fed to the query network. The query network outputs the predicted mean and standard deviation for the queried input $(x_{f^{i}}^q)$. 

\subsection{No-Reward Meta Learning}

 Provided that the change in dynamics can be represented in $\left(\boldsymbol{s}_{t}, \boldsymbol{a}_{t}, \boldsymbol{s}_{t+1}\right)$, NORML learns an pseudo-advantage function $A_{\psi}\left(\boldsymbol{s}_{t}, \boldsymbol{a}_{t}, \boldsymbol{s}_{t+1}\right)$. It is important to note that the aim of $A_{\psi}$ is to guide the meta-policy adaptation instead of fitting to the advantage function. $A_{\psi}$ is used to compute task specific parameters in the MAML inner loop from a set of state transitions of task i denoted as $D_{i}^{\text {train }}$ that does not contain reward signal. The learned advantage function is optimized in the MAML outer loop using the reward information present in rollouts ($D_{i}^{\text {test}}$) obtained from the updated task-specific policy.

\section{Proposed Method: Meta-World Conditional Neural Process (MW-CNP)}
In this section, we present our method, named Meta-World Conditional Neural Processes (MW-CNP), for learning world models using prior experience. We first describe the problem setup and then explain MW-CNP's structure in detail. 

In few-shot learning, the goal is to quickly adapt to an unseen target task using a few labeled data in the target environment. Meta-World Conditional Neural Processes (MW-CNP) can reduce the number of samples required from the target environment by generating world models from fewer samples from the target environment without accessing the target environment parameter. These models can then be used to obtain inexpensive rollouts for finetuning at test time. 

\paragraph{Online Meta-Learning}

We denote the initial state distribution as $\rho (s_0)  : \mathcal { S } \rightarrow \mathbb { R }$, state transition distribution of $task_i$  as ${\rho}_i { (s_{t+1}|s_t,a_t)} : \mathcal { S } \times \mathcal { A } \times \mathcal { S } \rightarrow \mathbb { R } $, and the reward function as $r_t : \mathcal { S } \times \mathcal { A } \times \mathcal { S }  \rightarrow \mathbb { R } $. During meta-training we store transitions for each $task_i$ as a set of observations $B_i=\left\{\left(s_{t}, a_{t},s_{t+1}\right)\right\}_{t=0}^{n} \subset S \times A \times S$ without the task parameter. It is important to note that during both training and testing the parameters of the state transition distribution are hidden. 

After we obtain the meta-policy and learn the pseudo-advantage function denoted by $A_{\psi}\left(\boldsymbol{s}_{t}, \boldsymbol{a}_{t}, \boldsymbol{s}_{t+1}\right)$ using NORML, we train our MW-CNP model using the replay buffer of transitions $B= \{B_{i}\}_{i=1}^{n}$ where n denotes the number of environments. 

\paragraph{Meta-World Conditional Neural Processes (MW-CNP)}

\begin{figure}[!htbp]
\includegraphics[width=\textwidth]{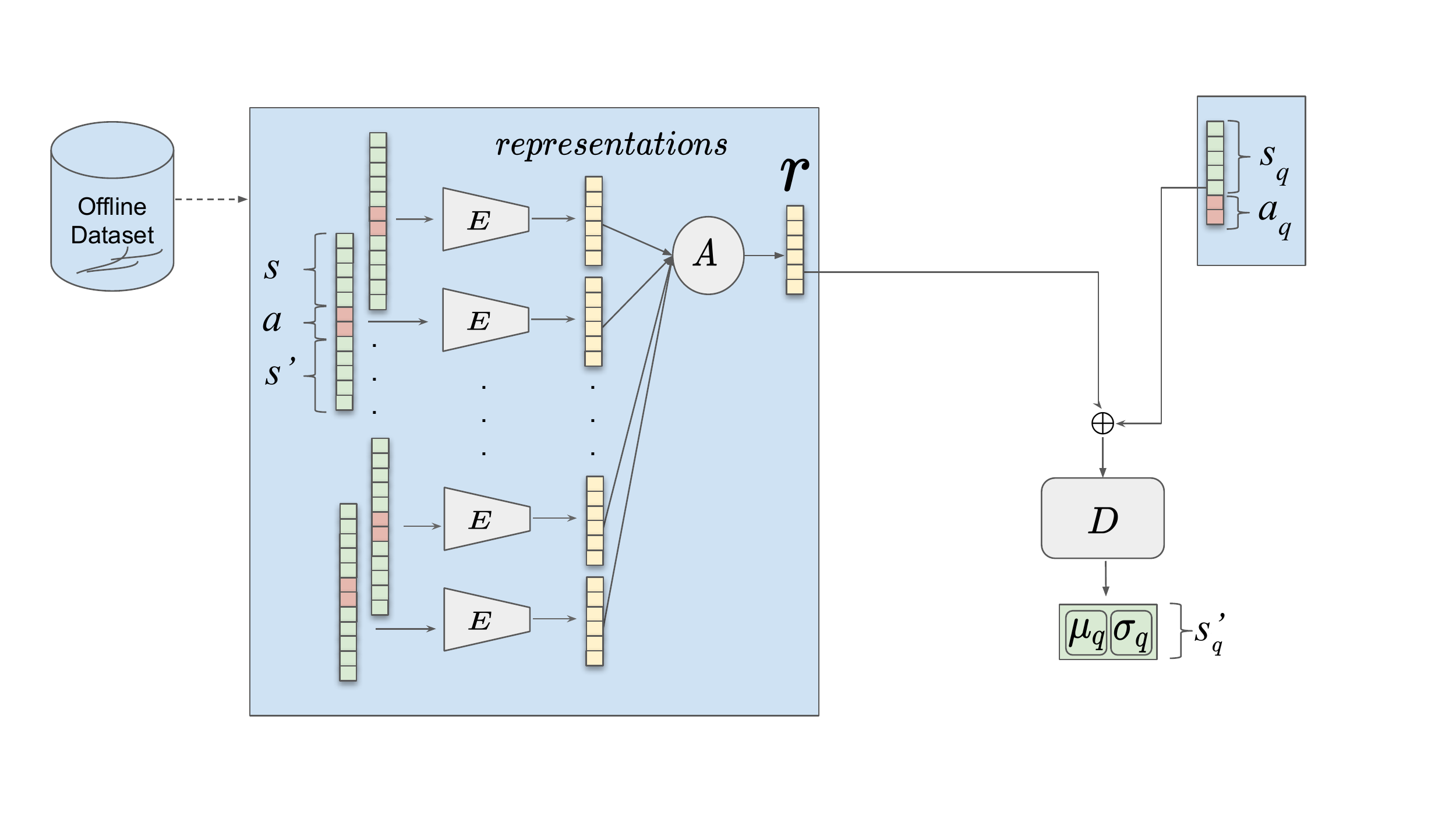}
\caption{Structure and the training procedure of the MW-CNP}
\label{fig:mwcnp}
\end{figure}

MW-CNP is trained in an offline fashion using unlabeled batches of Markov Decision Process (MDP) tuples collected during online meta-learning. Figure \ref{fig:mwcnp} illustrates the training procedure for MW-CNP. It is worth noting that the environment parameter is hidden during training and testing. In each training iteration, an unlabeled batch $B_i=\left\{\left(s_{t}, a_{t},s_{t+1}\right)\right\}_{t=0}^{n}$ is randomly sampled from the offline dataset. Then, a set of task-specific MDP tuples $\left\{\left(s_{k}, a_{k},s_{k+1}\right)\right\}_{k}$ and a single MDP $(s_q, a_q,s'_q)$ are randomly sampled from the chosen batch $B_i$. $(s_q, a_q,s'_q)$ is used for target state-action query $[s_q, a_q]$ and true target next state label $[s'_q]$.  Each MDP tuple $\left(s_{k}, a_{k},s_{k+1}\right)$ is encoded into a fixed size representation using a parameter sharing encoder network. These representations are then passed through an averaging module $A$ to obtain a latent representation $r$ of the hidden environment transition function used in batch $B_i$. The resulting latent representation $r$ is concatenated with the $[s_q, a_q]$ to predict the distribution parameters $\mu_q, \sigma_q$ of the next state $s'_q$ given the latent representation $r$ and the target query $[s_q, a_q]$. The loss function of MW-CNP can be expressed as

\begin{align*}
\mu_q, \sigma_q &= f_{\theta_{D}}\left([s_q, a_q] \oplus \frac{1}{n}\sum_{k}^{n} g_{\theta_{E}}\left(s_{k}, a_{k}, s'_{k}\right)\right)\\
\mathcal{L}(\theta_{E}, \theta_{D})&= -\log P\left(s_{q}^{'true} \mid \mu_q, \textit{softplus}(\sigma_q)\right)
\end{align*}

where $f_{\theta_{D}}$, $g_{\theta_{E}}$ are the decoder the encoder networks, $[s_q, a_q]$ is the target state action query, $\left(s_{k}, a_{k}, s'_{k}\right)$ are the randomly sampled transitions from the set of observations $B_i$. 

\begin{figure}[!htbp]
\includegraphics[width=\textwidth]{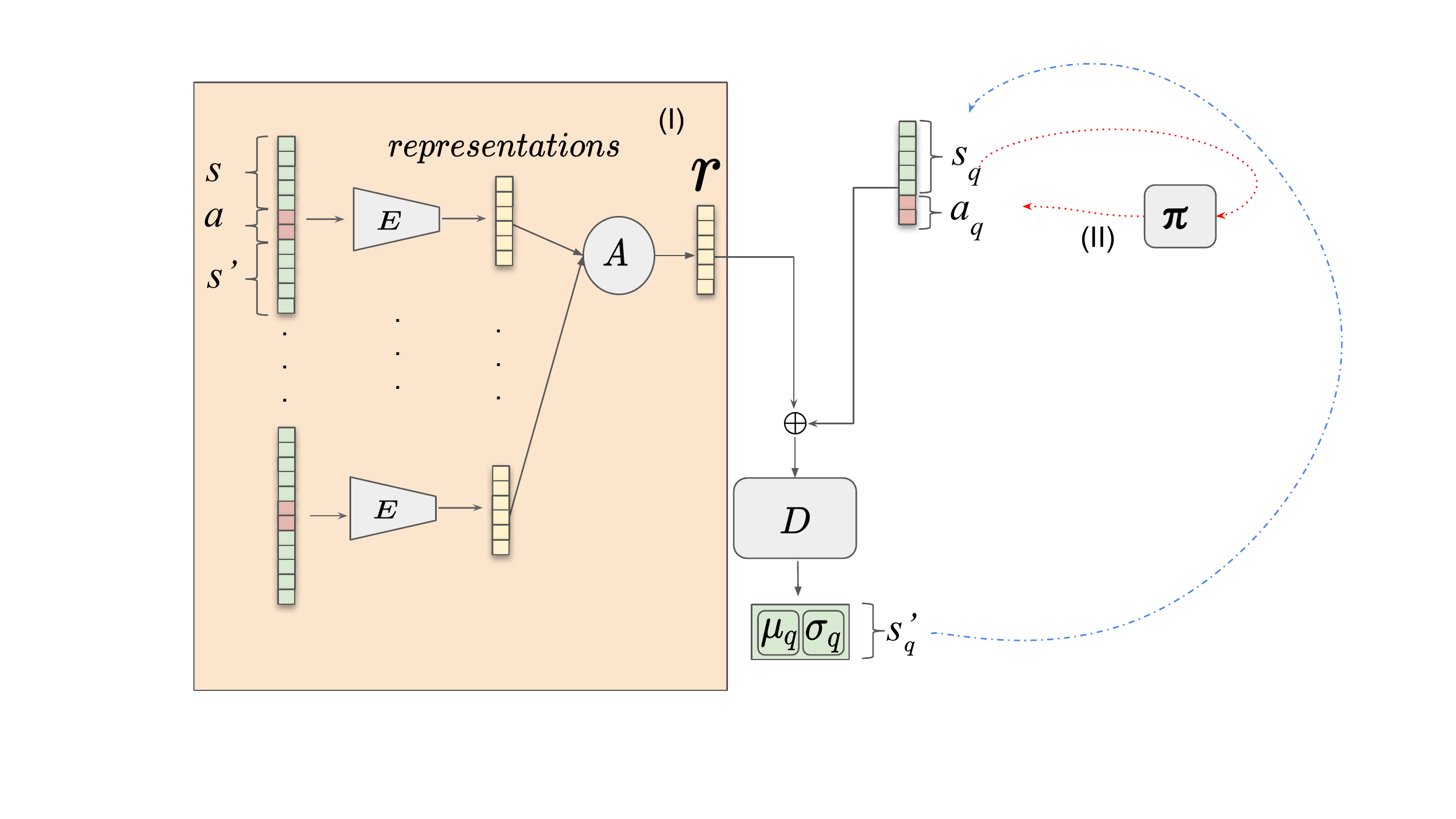}
\caption{Structure and the test procedure of the MW-CNP}
\label{fig:mwcnptest}
\end{figure}

Figure \ref{fig:mwcnptest} illustrates the test procedure of MW-CNP. At test-time, the agent is allowed to collect a few samples from the unseen target environment. These samples are encoded into representations of fixed size by an encoder network with shared weights. A shared representation of the target environment is obtained using an averaging module shown in Figure \ref{fig:mwcnptest}(I). Once the latent representation is obtained for the target environment with a hidden task parameter rollouts can be generated inexpensively. This representation is used to predict the parameters of the next-state distribution for the state-action query $[s_q, a_q]$. 

For each MW-CNP generated rollout, same true initial state, sampled from the real target environment, is used. The inital state is fed to the stochastic meta-policy to obtain the action query $a_q$ Figure \ref{fig:mwcnptest}(II). Then, the predicted next-state is used to sample the next action query until the episode is terminated. The rollouts "hallucinated" from the MW-CNP are combined with the true rollout sampled from the real target environment. The resulting set of rollouts are fed to the pseudo-advantage network. The learned pseudo-advantage network learned during online meta-learning uses $(s,a,s')$ tuple as input and outputs an advantage estimation value. Hence, once experience from the generated world model is collected, these experiences are fed to the learned advantage function in the form of $(s,a,s')$. Finally, meta policy is finetuned for fast adaptation to the target task using the estimated advantage values and combined set of MW-CNP generated rollouts and a single target environment rollout.

\section{Experiments}
In this section, we analyze the performance of our method and compare it with NORML and the oracle in 2D point agents, cartpole with sensor bias, and locomotion tasks. MW-CNP requires significantly less interaction with the target environment compared to NORML, for fast adaptation to the unseen task.

\subsection{2D Point Agent with Unknown Artificial Force Field}

The goal of the point agent, initialized at [x=0,y=0], is to move to the position [x=1,y=0], where x,y are the positions on the 2D plane. We are interested in a meta-RL setting used in \cite{yang2019norml} where the reward function is identical across multiple tasks. Different tasks are created by generating different artificial force fields that push the agent in different directions ($\phi$). We use the same reward function, the negative Euclidean distance from the goal position, and hyperparameters used in NORML for comparative analysis. In the Point Agent environment, 5000 tasks are defined over the $[-\pi,\pi]$ interval.

\begin{figure}[!htbp]
\includegraphics[width=\textwidth]{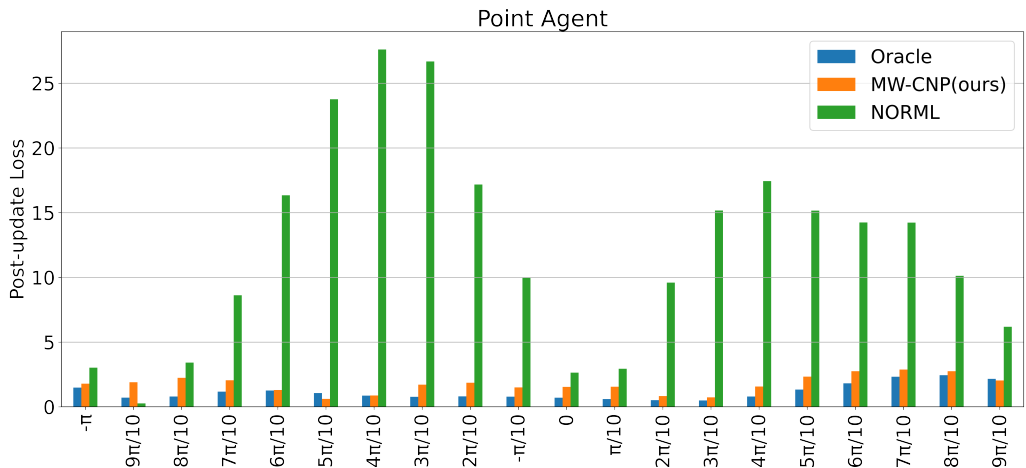}
\caption{Post-update loss (y-axis) of finetuning over generated data and ground truth data in target tasks. X-axis represents the target task indices.}
\label{fig:metacnp}
\end{figure}

The agent is initially trained across a distribution of 5000 tasks, i.e. in environments with 5000 different force fields. Then, it is tested in an unseen target task. The rollouts obtained from the target environment will be referred to as actual rollouts, whereas the rollouts generated from the MW-CNP will be named hallucinated rollouts. 

At the test time, Oracle agent uses 25 actual rollouts from the unseen target environment. 25 rollouts were used in the original NORML experiment; hence we use the same number of rollouts for comparison. The NORML agent and the MW-CNP use only a single rollout for fine-tuning the meta-policy. By limiting the number of actual rollouts that can be used for finetuning we aim to compare sample efficiency of MW-CNP to the baseline NORML. The average return obtained from the target environment after finetuning the meta policy of NORML with 1 rollout are shown in Fig. \ref{fig:metacnp} with green bars. As shown, when MW-CNP and NORML used the same amount of actual rollouts (1 actual rollout) from the target environment, MW-CNP outperformed NORML dramatically. Even though the oracle agent trained with NORML used a significantly higher number of actual rollouts sampled from the target environment than MW-CNP (25 to 1), their performances are similar, as shown in illustrations in Figures 4, 5, 6, and the bar plots in \ref{fig:metacnp}. 

Figure \ref{fig:metacnp} shows the expected post update return obtained in the target environment. We compare results for NORML finetuning over 1 rollout, NORML finetuning over 25 rollouts (Oracle), and finetuning over a combination of 24 hallucinated rollouts from the MW-CNP model and 1 actual rollout from the target environment. At test-time, MW-CNP uses a total of 25 mixed rollouts, similar to the oracle NORML, which uses 25 actual rollouts. For a reliable evaluation, we condition the MW-CNP model on the same rollout used for 1-rollout NORML fine-tuning. The results in Table \ref{table:pointtable} show that the samples generated from the agent's hallucination created by the MW-CNP can be used for finetuning the meta-policy for fast domain adaptation, thereby significantly increasing the sample efficiency in meta-testing.

\begin{table}[!htbp]
	\vskip\baselineskip 
	\caption[Point Environment]{2D Point Agent Environment Post Update Loss}
	\begin{center}
		\begin{tabular}{|c|c|c|}\hline
                 & mean &median \\\hline    
                MWCNP(ours)&1.74 $\pm 0.65$ & 1.75\\\hline
			NORML & 12.23 $\pm 7.81$ & 12.17 \\\hline    
			ORACLE  & 1.15 $\pm 0.59$ &0.84 \\\hline
		\end{tabular}
	\end{center}
	\label{table:pointtable}
\end{table}

\begin{figure}[!htbp]
\includegraphics[width=1\textwidth]{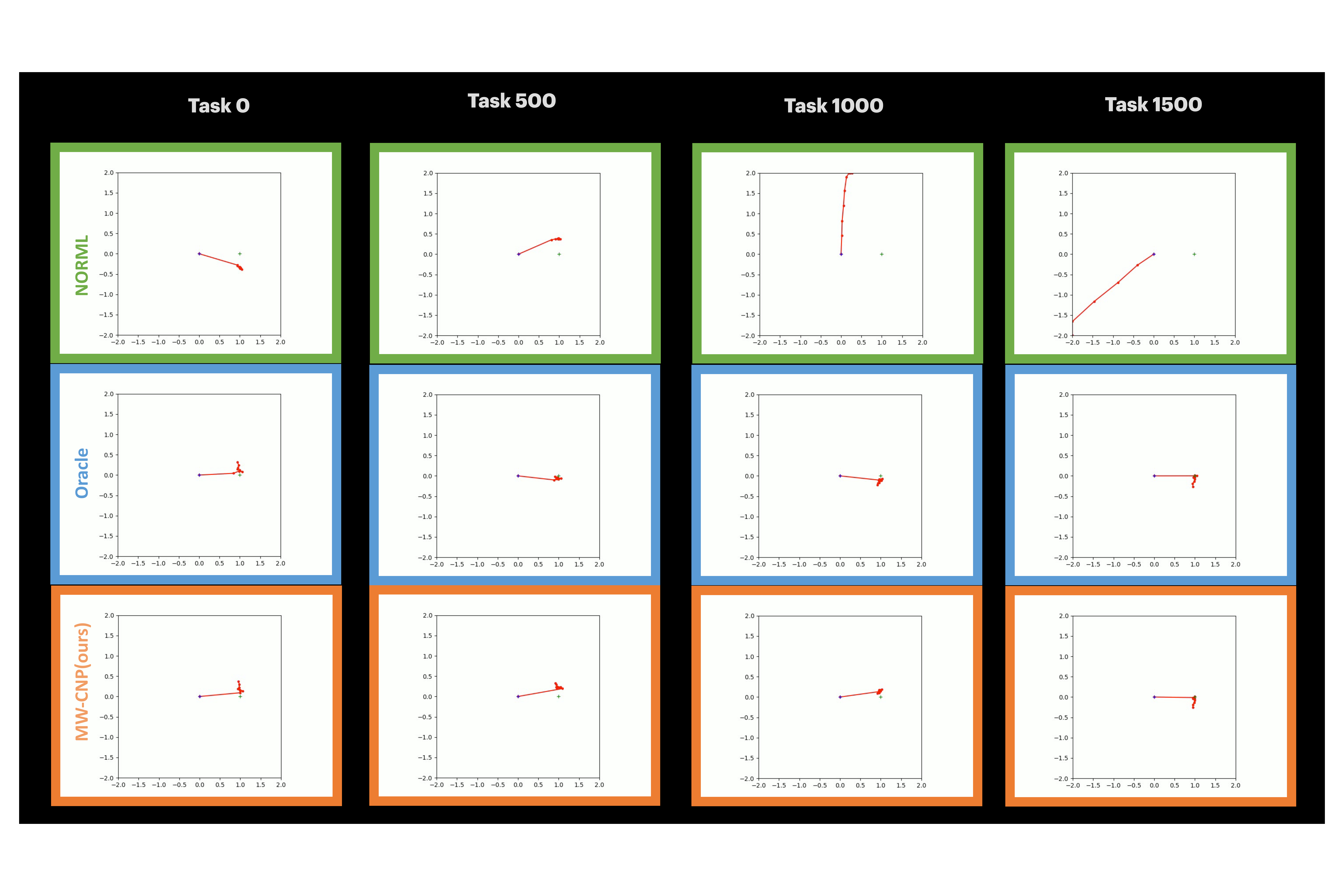}
\caption{Illustrations of trajectories for Tasks 0, 500, 1000 and 1500}
\label{fig:task0}
\end{figure}

\begin{figure}[!htbp]
\includegraphics[width=1\textwidth]{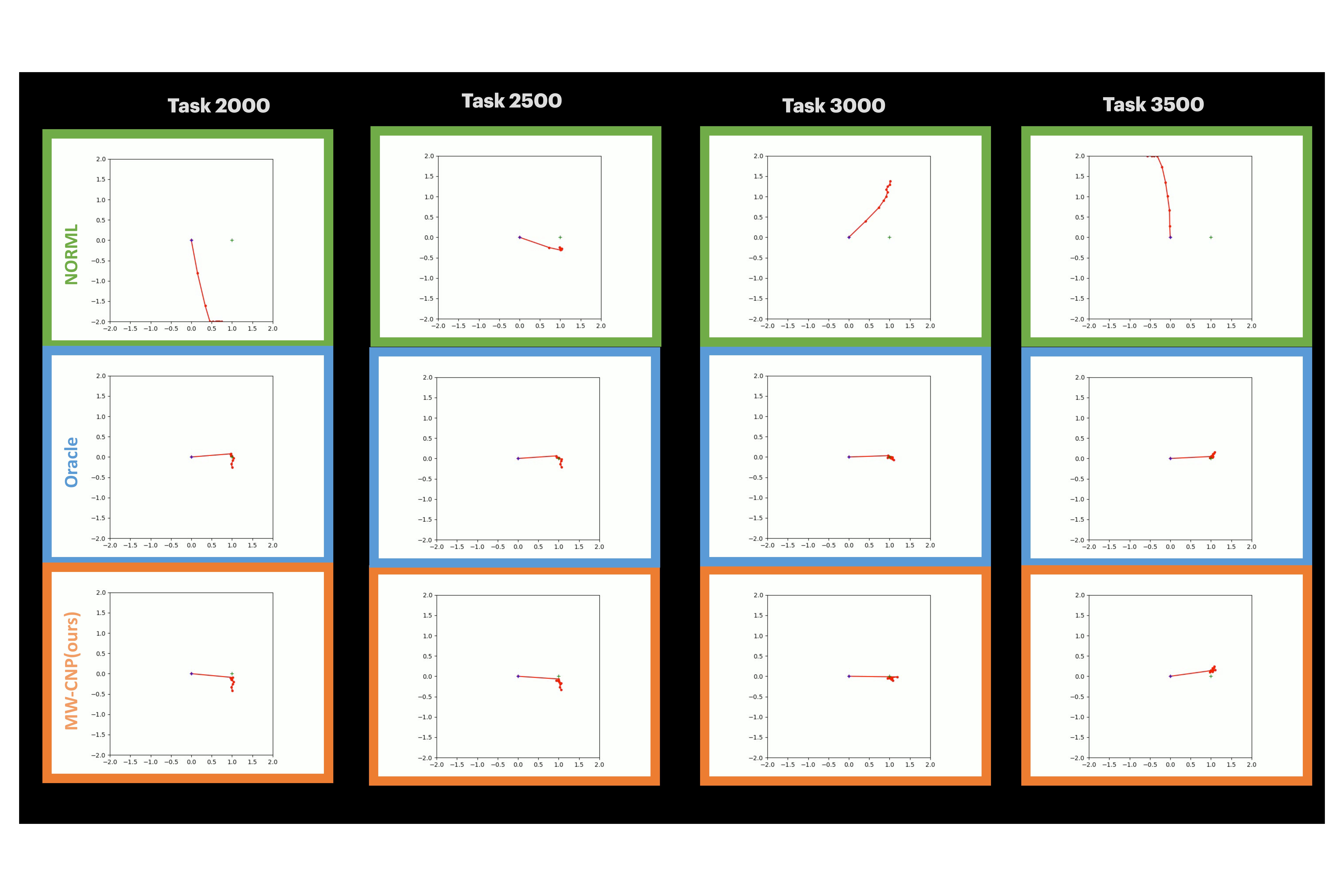}
\caption{Illustrations of trajectories for Tasks 2000, 2500, 3000 and 3500}
\label{fig:task1}
\end{figure}

\begin{figure}[!htbp]
\includegraphics[width=1\textwidth]{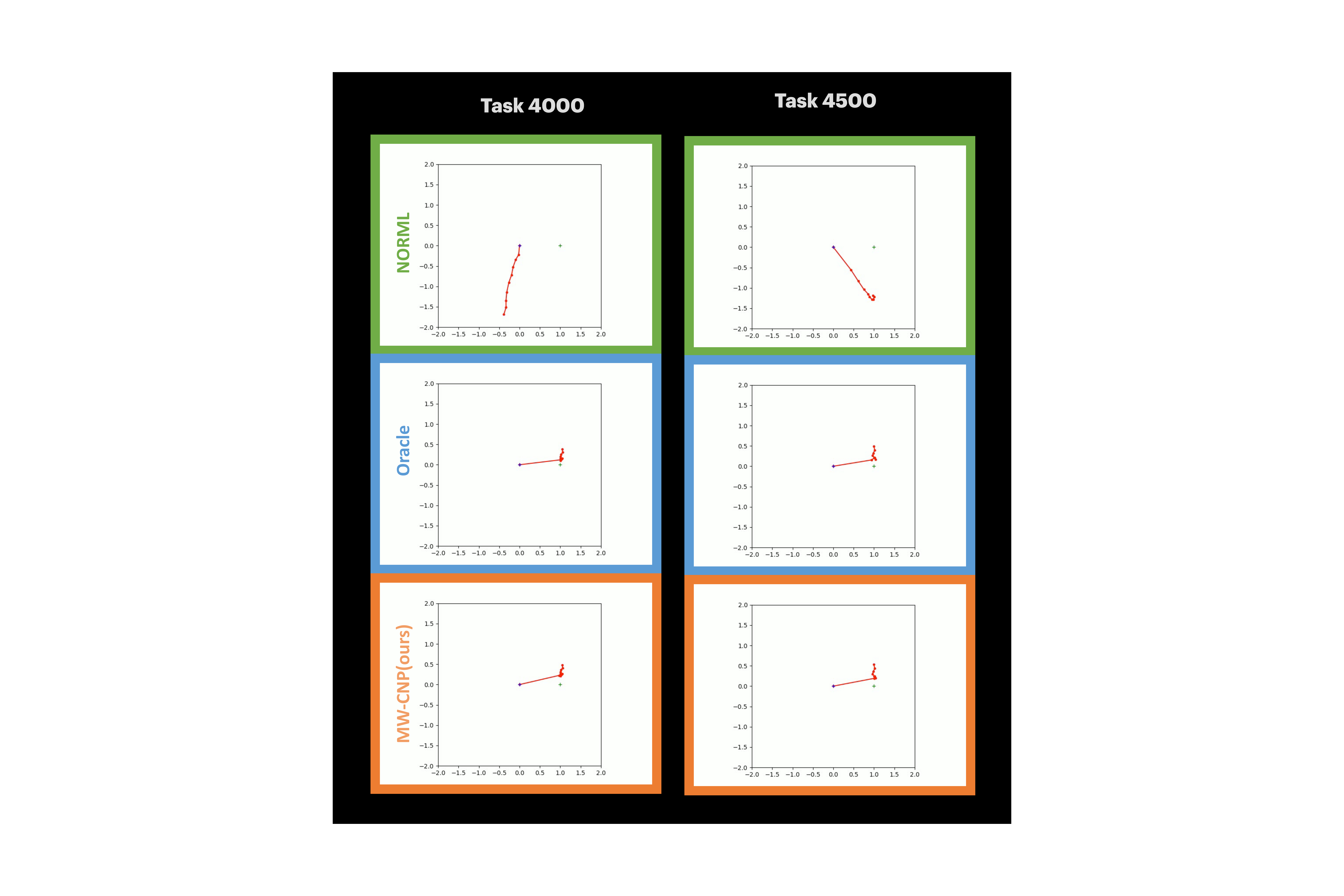}
\caption{Illustrations of trajectories for Tasks 4000 and 4500}
\label{fig:task2}
\end{figure}

\subsection{Cartpole with Unknown Angle Sensor Bias}

\begin{figure}[!htbp]
	\centering
		\subfigure[Cartpole with Unknown Angle Sensor Bias Environment]{
		\includegraphics[width=.46\textwidth]{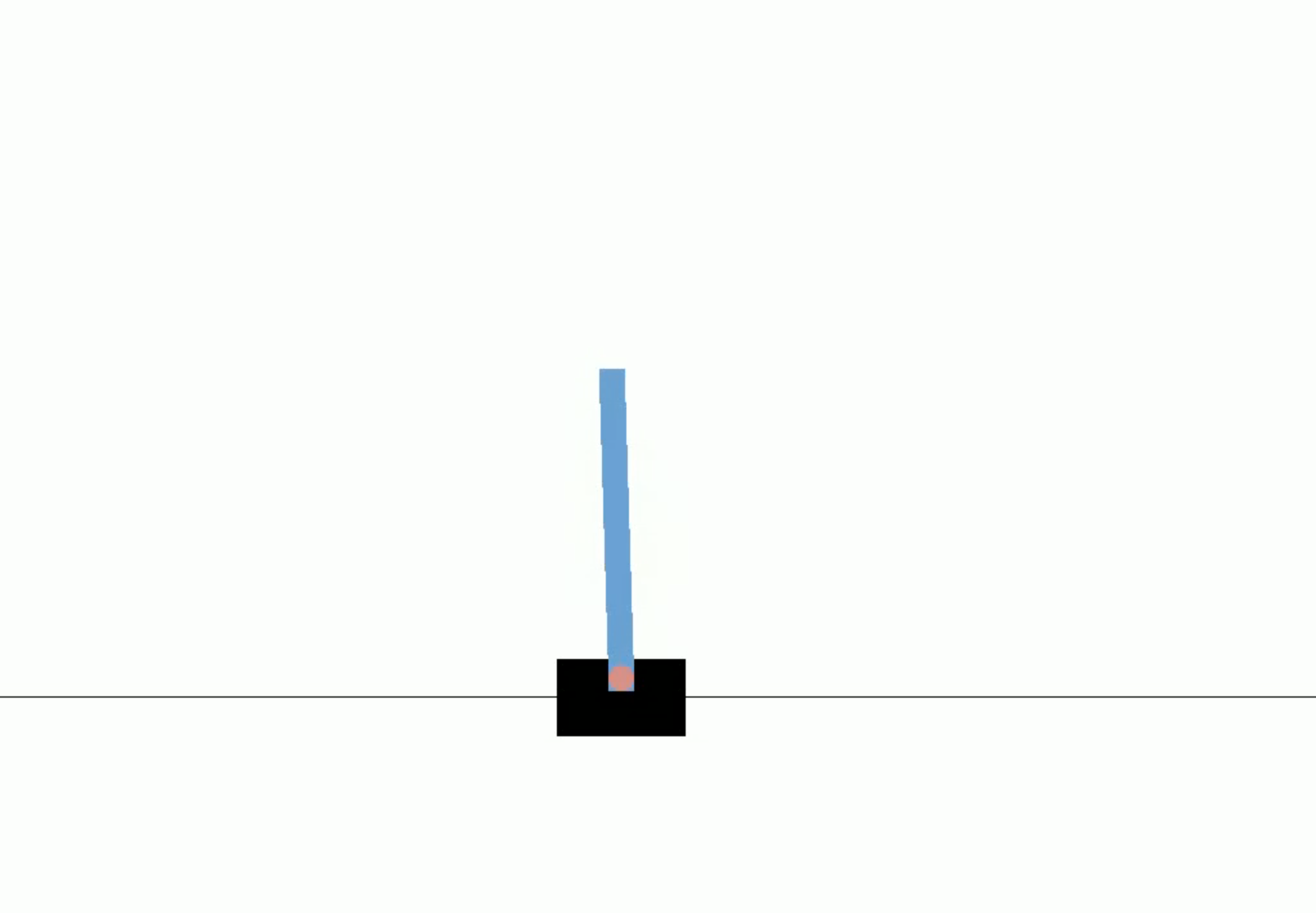}
	}
		\subfigure[Walker-2D Randomized Agent Dynamics Parameters Environment]{
		\includegraphics[width=.4\textwidth]{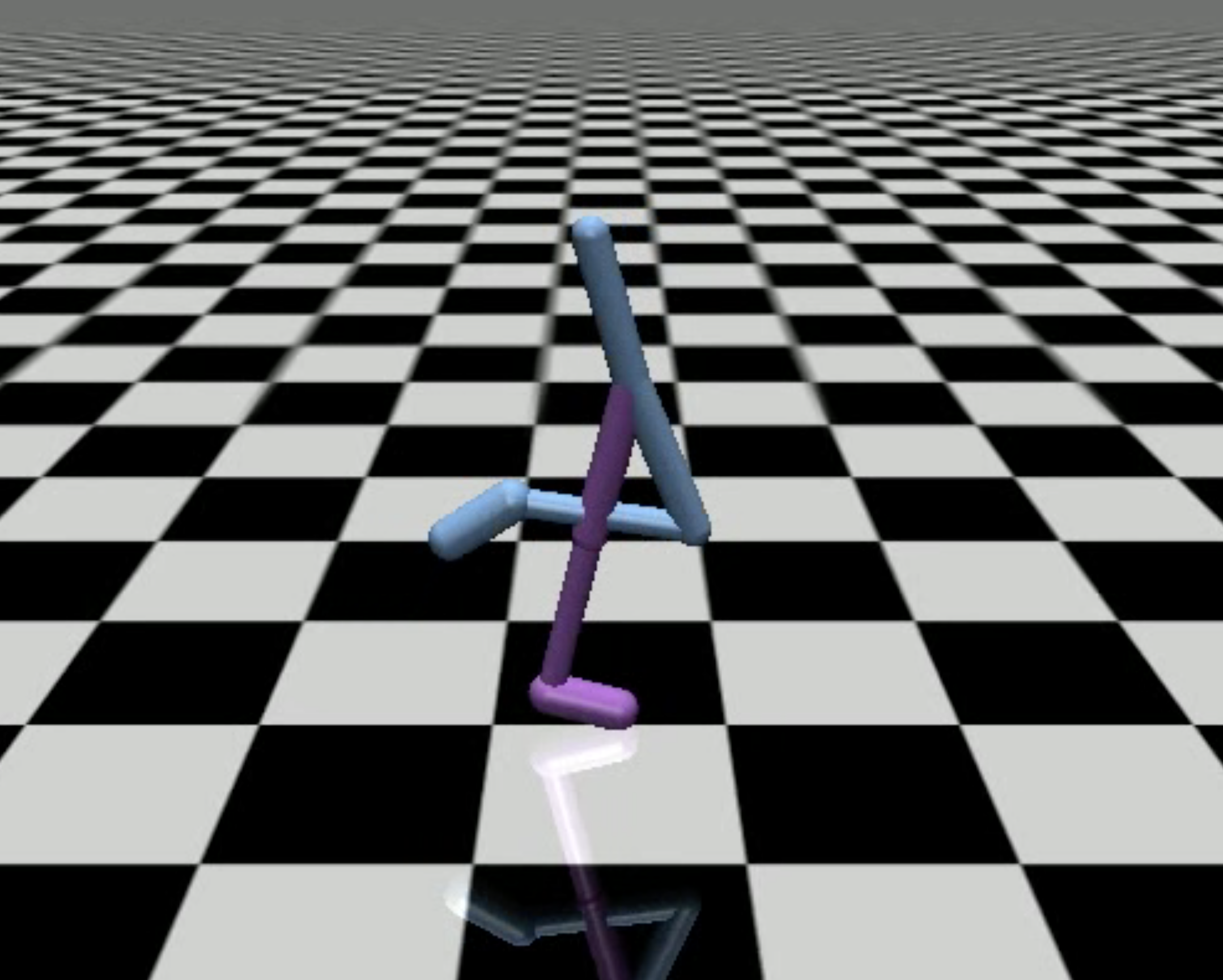}
	}
	\caption{Visualization of meta-test task performance in (a)cartpole , (b)walker environments}
	\label{fig:wenv}
\end{figure}

\begin{figure}[!htbp]
	\centering
		\subfigure[MWCNP obtained a significantly higher expected post-update reward of $386.44 \pm 159.27$ compared to NORML ($110.11 \pm 138.97$) using only 5 MDP tuples from the meta-test tasks.]{
		\includegraphics[width=.7\textwidth]{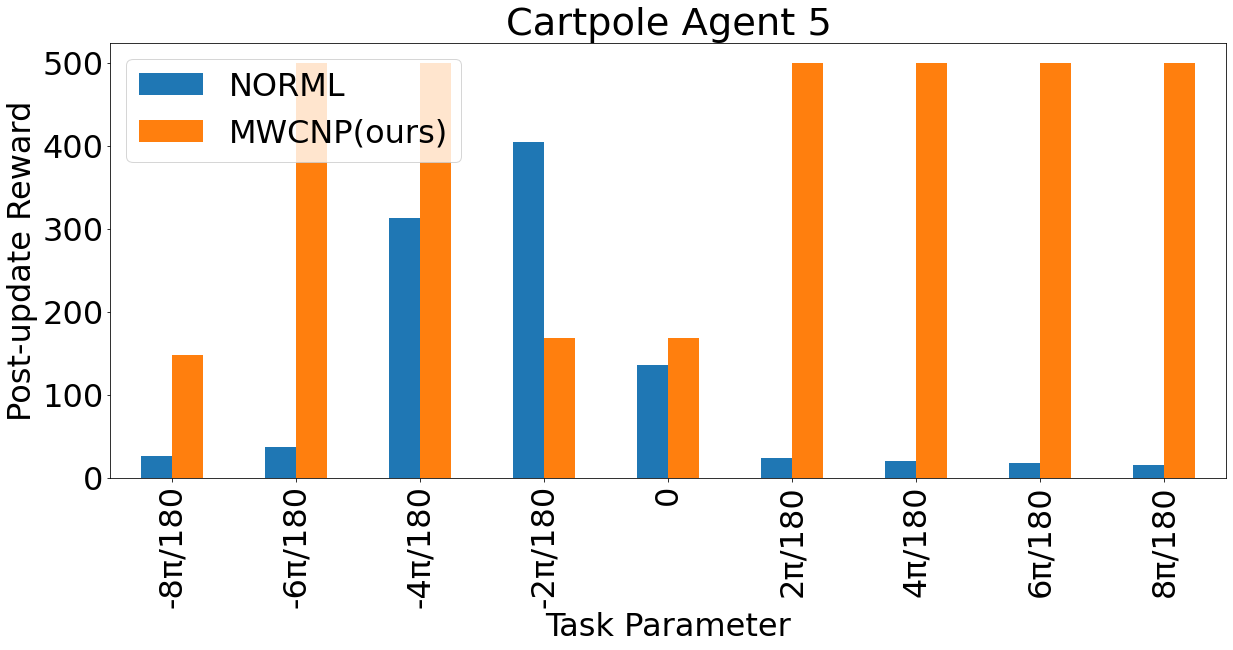}
	}
		\subfigure[Expected post update reward for MWCNP is $447.67 \pm 113.97$ using only 50 MDP tuples from the meta-test tasks. In contrast, NORML obtained an expected post-update reward of  $302.33 \pm 159.27$, which is still lower than the expected post-update reward of MWCNP finetuned over 5 samples. ]{
		\includegraphics[width=.7\textwidth]{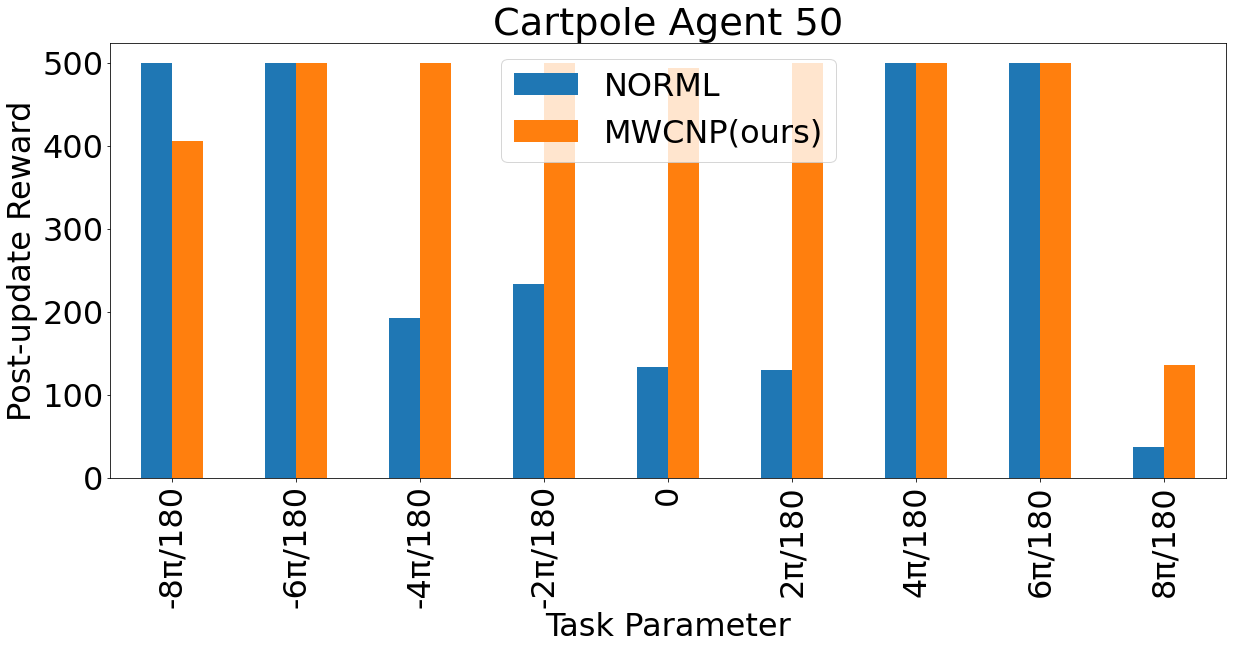}
	}
	\caption{Post-update reward (y-axis) of finetuning over generated data using (a) 5, (b) 50 target task MDP tuples and ground truth data in target tasks. X-axis represents the target task parameter.}
	\label{fig:cartpoleexp}
\end{figure}

We evaluate our method on the Cartpole with Angle Sensor Bias environment  visualized in Figure \ref{fig:wenv}a \cite{yang2019norml}. Tasks differ by a hidden position sensor drift in the range $[-8^{\circ},8^{\circ}]$. In meta-training, tasks are sampled from a uniform distribution where the transition dynamics function is parameterized by the aforementioned hidden variable. The state space consists of the position and velocity of the cart and the pole. We empirically found that the rollout length of 1000, used in prior work \cite{yang2019norml}, is not needed for adaptation in meta-testing. Hence to evaluate adaptation performance where a limited number of transition tuples are available, we compared the performance of MW-CNP to NORML with roll-out lengths of 5 and 50 shown in Figure \ref{fig:cartpoleexp}. Notice that the results are not symmetric across meta-tasks. This can be attributed to gradient bias in meta-training \cite{liu2022a}. Subsequently, in Figure \ref{fig:cartpoleexp}a we observe that this becomes prominent when evaluated with only 5 meta-test samples. Despite that, the meta-policy can successfully adapt to most test tasks using generated samples from the MW-CNP model.

\subsection{Walker-2D Randomized Agent Dynamics Parameters}
In order to explore the sample efficiency of MW-CNP in higher dimensional continuous control tasks, we evaluate our method in the Walker-2D Randomized Agent Dynamics Parameters environment (Walker-2D-Rand-Params)\cite{clavera2018learning} on MuJoCo\cite{todorov2012mujoco} visualized in Figure \ref{fig:wenv}b. Different locomotion tasks are created by unknown body mass, body inertia, degrees-of-freedom damping, and friction parameters. More specifically, the uniform distribution range used for sampling the latent scaling parameters is $[1.5^-3,1.5^3]$ for body mass, body inertia, friction parameters and $[1.3^-3,1.3^3]$ for body inertia.

\begin{figure}[!htbp]
\includegraphics[width=1\textwidth]{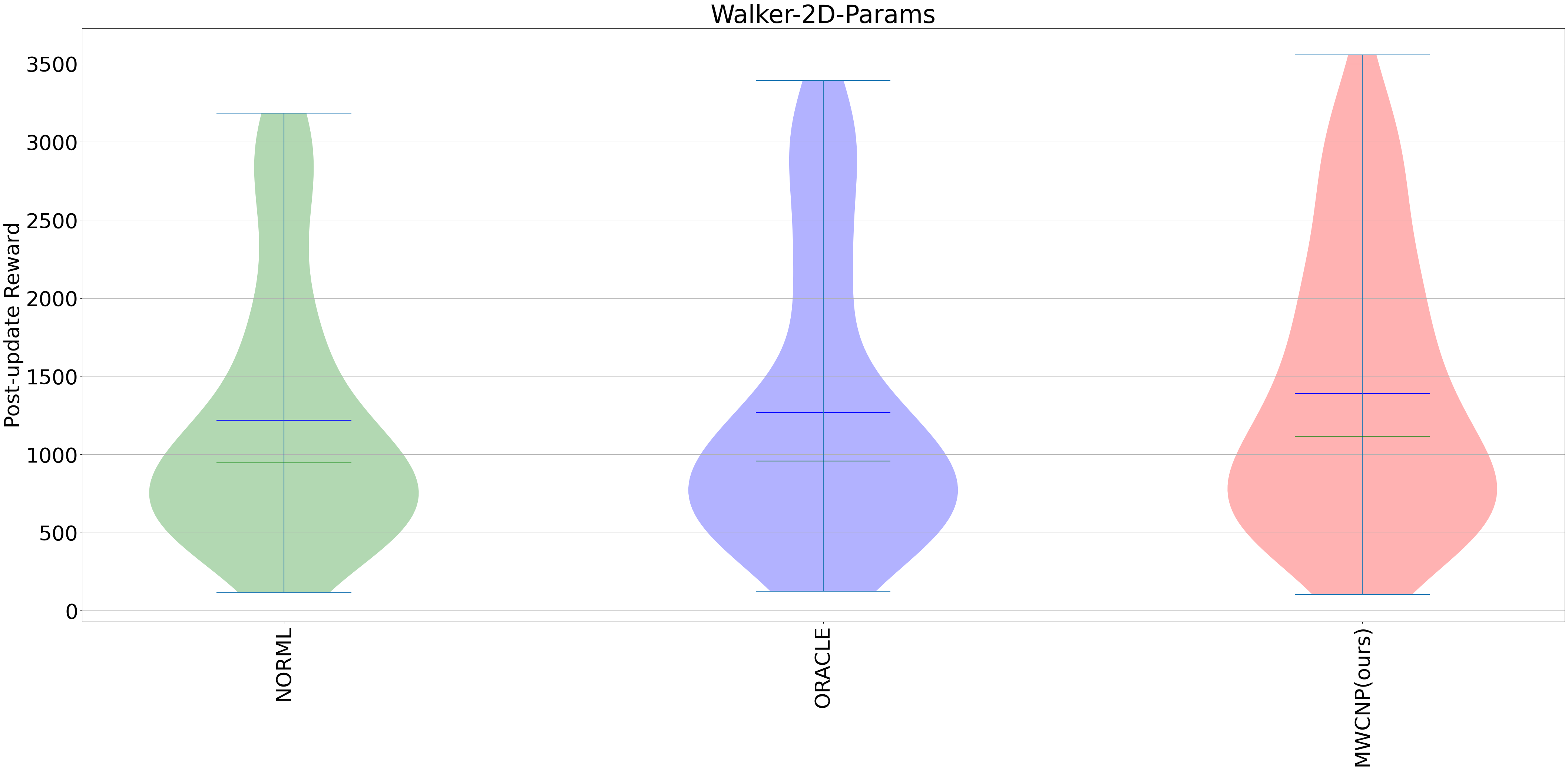}
\caption{Violin plots of post-update reward from 100 meta-test tasks. The violin plots illustrate the data's vertically aligned kernel density plots with clear markers indicating the minimum, maximum, mean, and median. The mean and median are represented by the blue and green lines, respectively.}
\label{fig:walkerexp}
\end{figure}

\begin{table}[!htbp]
	\vskip\baselineskip 
	\caption[Walker-2D-Params]{Walker-2D-Params  Post Update Reward}
	\begin{center}
		\begin{tabular}{|c|c|c|}\hline
                 & mean &median \\\hline   
                MWCNP(ours) & 1389.43 $\pm 862.32$& 1115.65\\\hline
			NORML & 1217.93 $\pm 810.48$ &944.86 \\\hline    
			ORACLE & 1268.46 $\pm 878.26$ &956.25 \\\hline
		\end{tabular}
	\end{center}
	\label{table:walkerresults}
\end{table} 

We sample 40 tasks for meta-training from a uniform distribution and use 100 unseen tasks for meta-testing. Figure \ref{fig:walkerexp} shows the post-update reward in meta-testing, where NORML and MW-CNP use a single rollout, whereas the ORACLE uses 25 rollouts for adaptation. We empirically demonstrate in Table \ref{table:walkerresults} that sample efficiency and meta-test adaptation performance have increased for a large range of unseen tasks.

\section{Conclusion}
We showed that meaningful hallucinated rollouts can be collected using the MW-CNP framework to guide the meta-policy adaptation. We compared the average meta-test performance obtained from finetuning over generated data with MW-CNP and the real simulated environment. MW-CNP's performance closely matched with the ORACLE and in the more complex locomotion task MW-CNP outperformed the ORACLE. MW-CNP model was able to generate samples from a significantly less number of MDP tuples in all our experiments thereby increasing sample efficiency during meta-testing.

All in all, we showed that using generated data for meta-updates, not only significantly increases sample efficiency but can also yield superior performance than using actual data. Furthermore, CNPs were shown to perform well with high dimensional inputs like image data (\cite{cnmp}). Extending our work and applying it to high-dimensional sensorimotor spaces, such as manipulator robots that use RGB-D cameras is an interesting research direction we intend to explore.

\paragraph{Declaration of Competing Interest} 
The authors declare that they have no known competing financial interests or personal relationships that could have appeared to influence the work reported in this paper.

\paragraph{Acknowledgements} 
This work was supported by the Scientific and Technological Research Council of Turkey (TUBITAK, 118E923) and by the BAGEP Award of the Science Academy. The numerical calculations reported in this work were partially performed at TUBITAK ULAKBIM, High Performance and Grid Computing Center (TRUBA resources).

\bibliographystyle{unsrt}  
\bibliography{references}

\end{document}